\documentclass[11pt]{article}

\usepackage[utf8]{inputenc}
\usepackage[T1]{fontenc}
\usepackage[a4paper,margin=1in]{geometry}

\usepackage{graphicx}
\usepackage{comment}
\usepackage{caption}
\usepackage{subcaption}
\usepackage{array}
\usepackage{booktabs}
\usepackage{arydshln}
\usepackage{multirow}
\usepackage{float}
\usepackage{xcolor}
\usepackage{algorithm}
\usepackage{algorithmic}
\usepackage{amsmath}
\usepackage{amssymb}
\usepackage[hidelinks]{hyperref}
\usepackage{makecell}
\usepackage{enumitem}
\usepackage{footmisc}
\usepackage{tablefootnote}
\usepackage{pdfpages}
\usepackage{microtype}

\title{A Prototypical Signature Approach for Writer-Independent Offline Signature Verification}

\author{
Kecia G. de Moura, Robert Sabourin, Rafael M. O. Cruz \\[0.7em]
École de technologie supérieure -- Université du Québec \\
Montreal, Québec, Canada \\[0.3em]
\texttt{kecia.gomes-de-moura.1@ens.etsmtl.ca} \\
\texttt{\{rafael.menelau-cruz,robert.sabourin\}@etsmtl.ca}
}

\date{}

\begin{document}

\maketitle

\begin{abstract}
Offline handwritten signature verification aims to distinguish genuine from forged signatures using static images. Since real forgeries are rarely available, negative samples are usually randomly drawn from genuine signatures of other users to create training data. However, this random selection often lacks diversity, increases redundancy, and escalates computational cost, leading to inefficient training. We propose a data-driven strategy to generate diverse, informative negative samples using prototypical signatures, which are compact, non-identifiable summaries of genuine signature features. Based on the experiments results, we conclude that (i) prototypical signatures yield more informative negative samples, improving the detection of skilled forgeries; (ii) the proposed approach is backbone-agnostic showing robustness across architectures; and (iii) when combined with a primal-form linear SVM, it serves as an alternative to RBF-based models while significantly improving scalability and computational
efficiency. Implementation of the method is available at
\url{https://github.com/kdmoura/proto_hsv.}
 
\vspace{0.5em}
\noindent\textbf{Keywords:} Offline handwritten signatures, Data summarization, Prototype generation, Writer-independent system, Biometrics, Scalability
\end{abstract}

\section{Introduction}  
\label{sec:introduction}  

Handwritten Signature Verification (HSV) systems are biometric authenticators that detect which signatures belong to a claimed individual and which are created by an impostor. Generally, such detectors are categorized into online and offline systems \cite{Hameed2021}. Offline HSVs focus on static signature images acquired when the signing process is finalized, in contrast to online systems that exploit dynamic features during the writing process \cite{Hameed2021}.

Offline HSV systems can implement two approaches: writer-dependent (WD) and writer-independent (WI). In the WD approach, each enrolled user has their own dedicated classifier, which returns higher verification performance but increases complexity and maintenance, as it requires individual training data and a model for each new user. In contrast, WI systems utilize a single classifier for all users, reducing complexity and improving generalization at the expense of a lowered performance \cite{Singla2025_HSV_survey}.

In HSV, a signature can be classified into \textit{genuine}, produced by the owner, and \textit{forgery}, created by someone else.  Systems are particularly interested in detecting skilled forgery signatures, as they are simulations of the original sample, making them difficult to distinguish from the original \cite{Singla2025_HSV_survey}. However, these samples are not always available during the training stage  \cite{Diaz2019_perspective}, forcing systems to rely exclusively on genuine signatures to learn models that should recognize forged signatures without directly modeling their patterns. As an alternative, a widely adopted approach is to use random forgeries, which are genuine signatures from other users \cite{LI2024_TransOSV,Prajapati2021_dataAugmentationFeatureSpace,Tsourounis2022_singleSignature,Vasilakis2025_RiemannianDichotomizer_onSPD,Zhang2024_RF_HSV_Disentangling}. 

\begin{figure*}[!h]
    \centering
\includegraphics[
trim=0 25 0 0, clip,
    width=\linewidth]{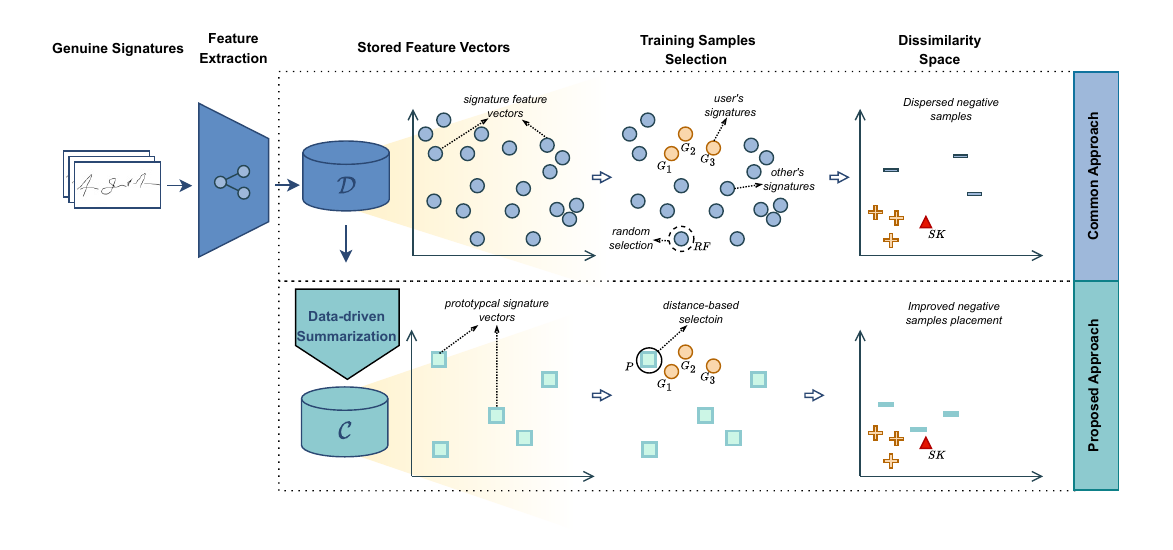}
    \caption{Illustration of the traditional HSV approach (top) versus the proposed method (bottom) for generating training data. In the conventional approach, feature vectors extracted from signature images are stored in the development set ($\mathcal{D}$), from which random forgeries ($RF$) are sampled independently of genuine signatures ($G$), often resulting in dispersed negative samples. Our approach performs a data-driven summarization on $\mathcal{D}$ to generate a compact set of prototypical signature vectors stored in the summary set ($\mathcal{C}$). These vectors are selected based on their distances to genuine instances, producing negative samples that better approximate the characteristics of skilled forgeries ($SK$).}
    \label{fig:fig_prototype_vs_common}
\end{figure*}

In the context of writer-independent HSV, one approach is to train a single classifier on a dissimilarity dataset with positive and negative samples. Positive samples are obtained from comparisons between genuine signatures of the same user, and negative ones from genuine signatures of different users \cite{Talles2023_AmultitaskApproach4ContrastiveLearning}. This process is illustrated in Figure~\ref{fig:fig_prototype_vs_common}(top). Previous research \cite{Souza2019_CNN_Prototype_OnDissimilarRepresentation} indicates that such sampling introduces redundancy, increasing computational and storage costs. Moreover, while positive samples cluster near the origin (with low intra-class dissimilarity), random forgeries are widely scattered \cite{Souza2019_CharacterizationOfHSinDissimilarityRepresentationSpace}, thereby limiting their usefulness for detecting skilled forgeries.

The limitations of using random forgeries in writer-independent HSV highlight the need for more informative training samples to improve performance, efficiency, and scalability. Yet, identifying such samples, especially those resembling a target user’s signatures, can be computationally prohibitive, requiring exhaustive comparisons across all users. To overcome this, we propose a data-driven summarization approach for generating training data. It enables the selection of informative samples while decreasing redundancy, storage demands, and computational effort. A conceptual overview of the proposed method is presented in Figure~\ref{fig:fig_prototype_vs_common}(bottom). Our method summarizes the complete collection of signature feature vectors by dividing them into distinct clusters that group samples sharing similar traits. This clustering yields a compact set of \textit{prototypical signatures} that model the distribution of the development dataset. These representatives serve as candidates for the negative samples selection based on their distance from genuine user signatures. By adopting this distance-based strategy, the system can identify more challenging negative examples, ultimately enhancing the classifier’s capability to recognize skilled forgeries.

We evaluate the proposed approach in a writer-independent setting using three benchmark datasets: GPDS Synthetic, CEDAR, and MCYT-75. Two classifiers are tested, a Radial Basis Function (RBF) SVM and a Linear SVM optimized with Stochastic Gradient Descent (SGD). Results show that the method performs consistently across different backbone models and significantly reduces storage and computational costs when paired with a linear SVM, supporting scalable real-world deployment.

The main contributions of this work are as follows:  

\begin{itemize}[
topsep=0pt,
label=\textbullet,
]
    \item A method based on prototypical signatures for generating informative samples for 
    WI-HSV training.  

    \item A scalable and resource-efficient solution for writer-independent systems, reducing computational cost and enabling large-scale deployment.

    \item We demonstrate that our method is backbone-agnostic and integrates seamlessly with existing feature extractors, improving flexibility and reuse.
\end{itemize}

\section{Proposed Method}
\label{sec:proposed}

\begin{figure*}[!ht]
    \centering
\includegraphics[
    trim=0 15 140 0, clip,
    width=\linewidth]{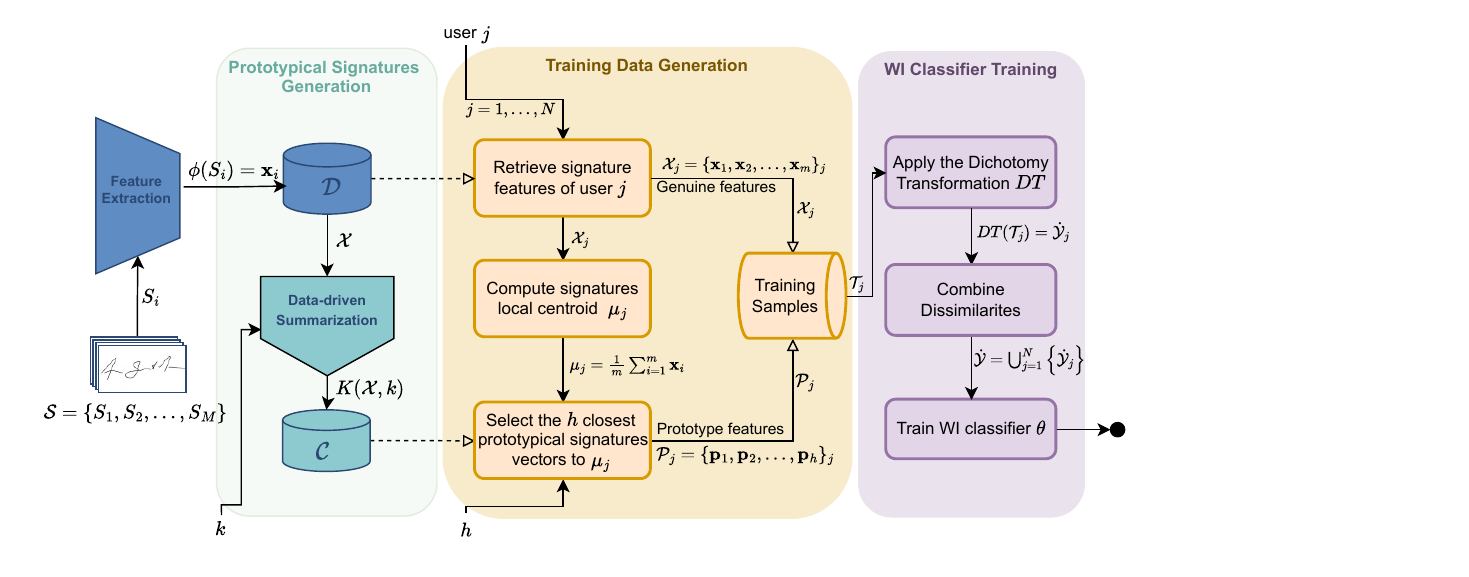}
    \caption{Overview of the proposed method. A set of handwritten signature images $\mathcal{S}$ is converted into feature vectors $\mathcal{X}$ via a feature extractor $\phi(\cdot)$ and stored in the development set $\mathcal{D}$. These vectors undergo a data-driven summarization process $K(\mathcal{X}, k)$ producing prototypical signatures that are stored in the summary set $\mathcal{C}$. For each user $j$, a local centroid $\boldsymbol{\mu}_j$ is computed from their genuine signature features $\mathcal{X}_j$. The $h$ closest prototypical signatures $\mathcal{P}_j \subset \mathcal{C}$ are then selected based on Euclidean distance to $\boldsymbol{\mu}_j$ and combined with genuine features to form the training set $\mathcal{T}_j$. A dichotomy transformation is applied to each $\mathcal{T}_j$, and the resulting set is employed to train a WI classifier.}
    \label{fig:fig_prototype_proposed}
\end{figure*}

Our method is a training data generation strategy designed to refine the selection of negative samples for writer-independent HSV systems. Figure~\ref{fig:fig_prototype_proposed} presents an overview of the proposed method\footnote{A list of symbols and notation is provided in \href{wi_proto_hsv_supp.pdf\#supp_symbol}{Section 2} of the supplementary material.}.

As shown in Figure~\ref{fig:fig_prototype_proposed}, the process starts with a set of \( M \) offline handwritten signature images, \( \mathcal{S} = \{S_1, S_2, \dots, S_M\} \), derived from \( N \) distinct users. A corresponding set of feature vectors \( \mathcal{X} = \{\mathbf{x}_1, \mathbf{x}_2, \dots, \mathbf{x}_M\} \), with \( \mathbf{x}_i \in \mathbb{R}^d \), where $d$ denotes the feature dimensionality, is extracted utilizing a representation model \( \phi(\cdot) \).

Afterward, the feature vectors in \( \mathcal{D} \) are submitted to a data-driven summarization process \( K(\mathcal{X}, k) \). In  \( K(,)\), vectors are grouped into \( k \) regions and prototypical signatures \( \mathcal{C} = \{\mathbf{c}_1, \mathbf{c}_2, \dots, \mathbf{c}_k\} \) are obtained. Vectors from $\mathcal{C}$ serve as candidate negative instances, whereas samples in \( \mathcal{D} \) form the positive class to construct the training set \(\mathcal{T} \).

\subsection{Prototypical Signatures Generation}

\begin{figure}[ht!]
    \centering
\begin{subfigure}{0.32\textwidth}
        \includegraphics[width=\linewidth]{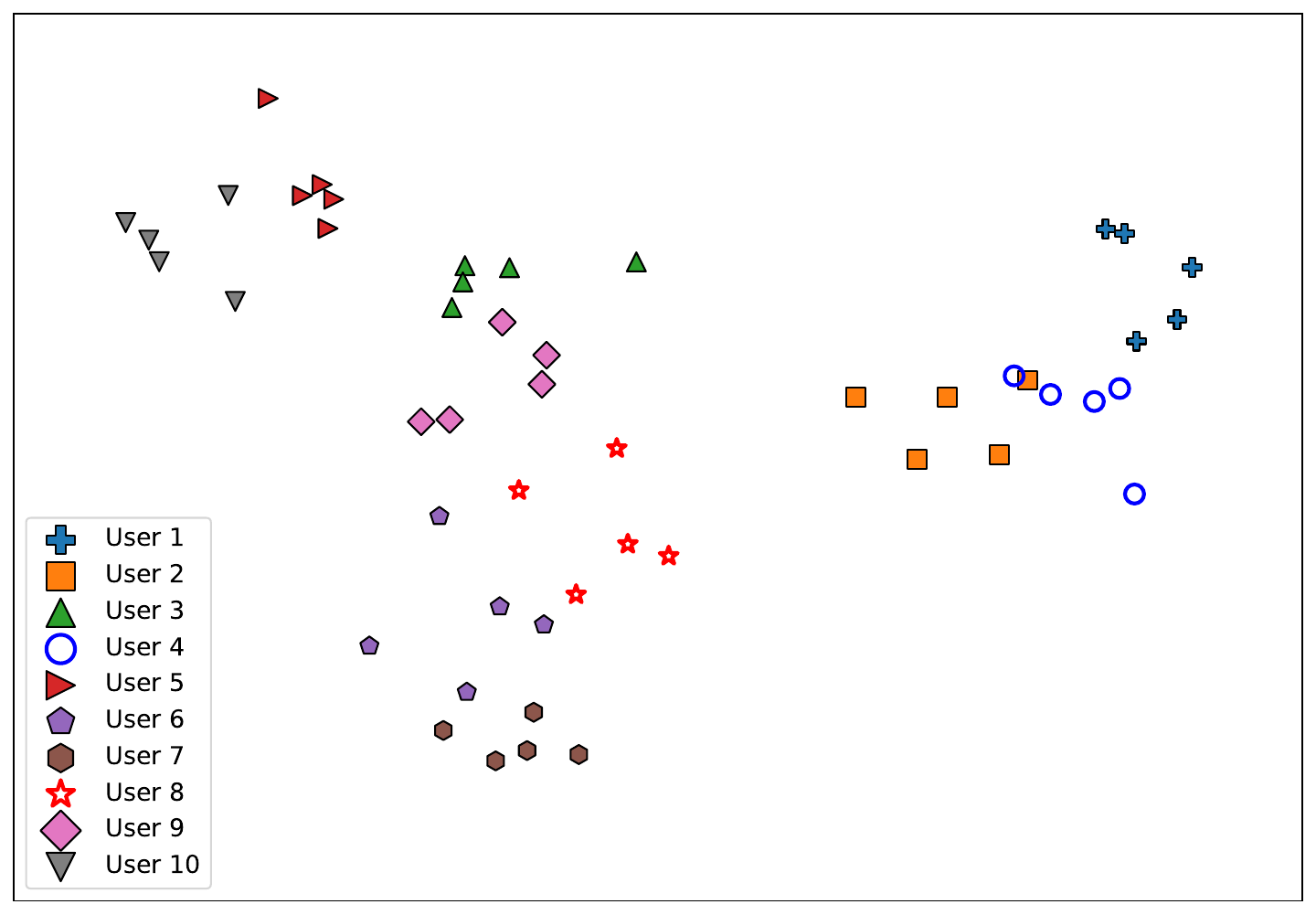}
        \caption{Signature vectors.}
        \label{fig:fig_example_a}
    \end{subfigure}
    \hfill
    \begin{subfigure}{0.32\textwidth}
        \includegraphics[width=\linewidth]{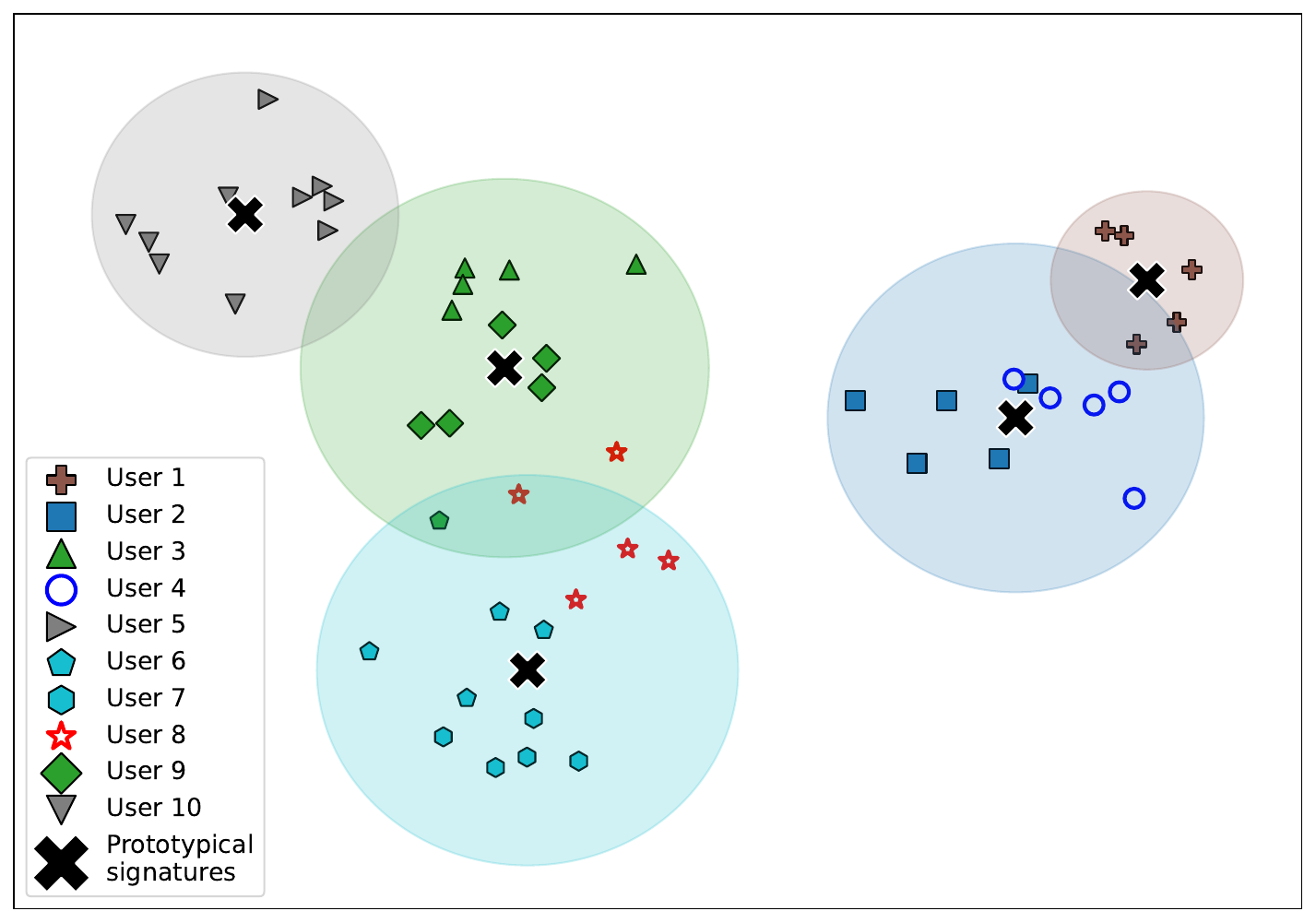}
        \caption{Summarization.}
         \label{fig:fig_example_b}
    \end{subfigure}
    \hfill
    \begin{subfigure}{0.32\textwidth}
        \includegraphics[width=\linewidth]{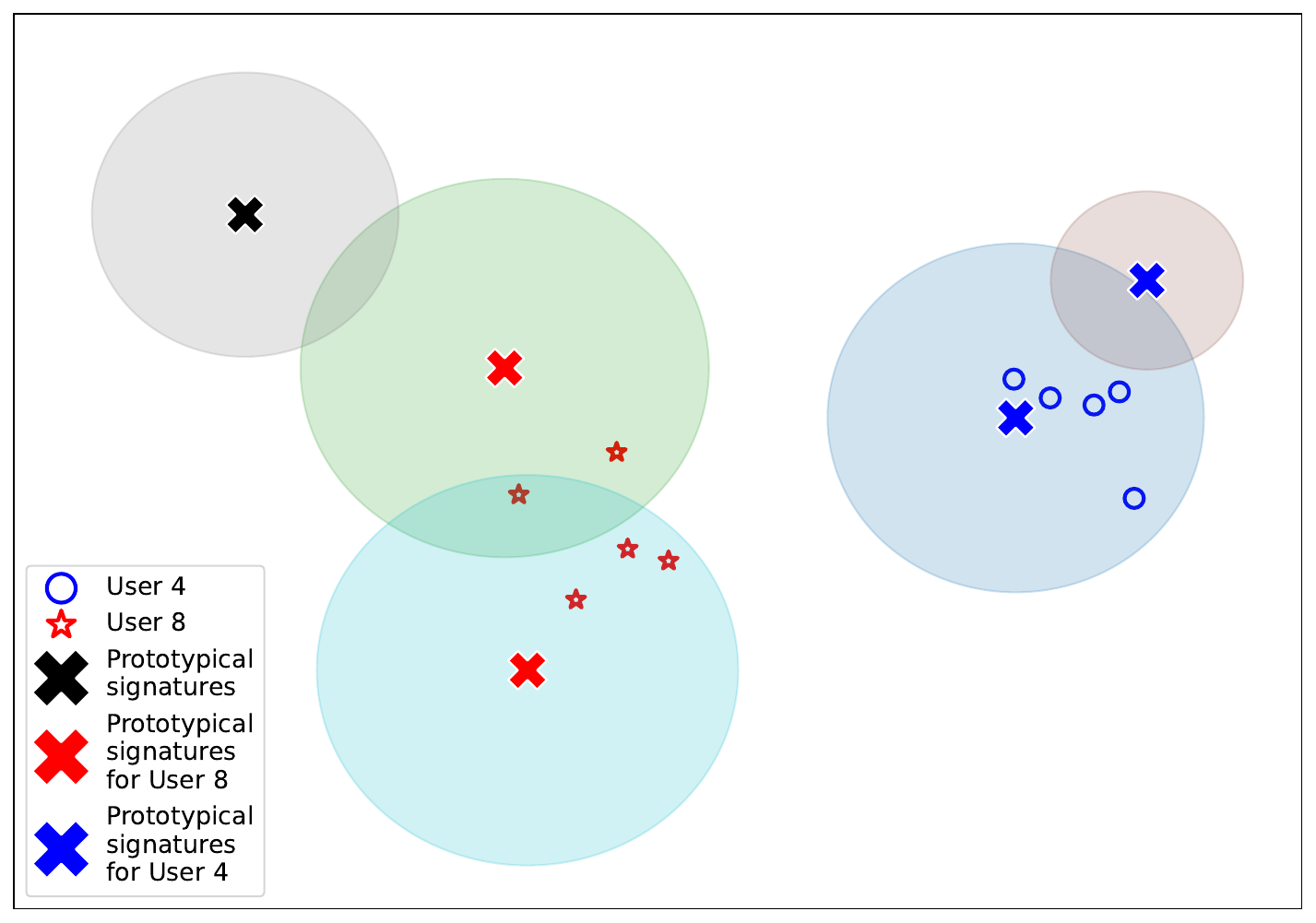}
        \caption{Sample selection.}
        \label{fig:fig_example_c}
    \end{subfigure}
    
    \caption{Prototypical signatures generation and selection on a toy sample. (a) Signature feature vectors from multiple users. (b) An example of summarization using $k=5$, reducing the feature space to a representative subset. (c) Distance-based selection of two negative samples for \textit{User 4} and \textit{User 8}.}
    \label{fig:fig_example}
\end{figure}

The main concept of the proposed approach is depicted in Figure~\ref{fig:fig_example}. To present the idea, we use a toy dataset containing $N=10$ users, each contributing 
$m=5$ handwritten signatures. In total, the dataset comprises $M = 50$ samples ($N \times m$), which are displayed in a two-dimensional feature space.  

Our goal is to select the most informative samples for training, specifically those that enhance the detection of skilled forgery signatures, which are considered the most challenging type of forgery due to their high visual similarity to genuine signatures~\cite{Singla2025_HSV_survey}. Ideally, one would compare each user's signatures with all others to identify the most similar ones through an exhaustive search. However, this approach,  poses two main challenges. First, the computational cost may become prohibitive for large datasets due to the sheer number of possible comparisons. Second, a strategy based solely on proximity can be detrimental, as it can filter out diverse samples that are vital for robust training \cite{roh2021sample}.

To address these problems, the summarization process (Figure~\ref{fig:fig_example_b}), considerably narrows the search space by finding \emph{prototypical signature vectors} that act as potential forgery candidates. The regions associated with these vectors contain signatures of multiple users, enabling them to encompass characteristics of a diverse number of samples. \textbf{We hypothesize that selecting prototypical vectors that lie closest to users' genuine signature samples can increase the discriminative capacity of the classifier's decision boundaries}. 

The use of prototypical signatures facilitates the efficient selection of challenging negative samples, while simultaneously mitigating the risk of over-reliance on proximity (Figure~\ref{fig:fig_example_c}). In terms of computational burden, with $k$ partitions, the proposed method requires approximately $\frac{N \times m}{k}$ times fewer comparisons than would be needed without data-driven summarization. 

The summarization step is implemented via clustering over all genuine signature features in the development set. In this work, we adopt the \( k \)-means algorithm due to its scalability, simplicity, and ability to produce meaningful centroids \cite{Rada2023_kmeans_just}. The resulting centroids constitute the set \( \mathcal{C} = \{\mathbf{c}_1, \mathbf{c}_2, \dots, \mathbf{c}_k\} \), forming a compact, non-identifiable representation of the population's signature distribution.

\subsection{Training Data Generation}
As presented in Figure~\ref{fig:fig_prototype_proposed} and detailed in Algorithm~\ref{algo:data_generation}, the training data generation process comprises three steps that are executed for each user: (1) retrieval of signature feature vectors, (2) computation of the local centroid based on all genuine signatures, and (3) selection of the closest prototypical signatures in relation to the local centroid. Formally, let $m$ be the number of signatures contributed by each user, and  \(\mathcal{X}_j =  \{ \mathbf{x}_{1}, \mathbf{x}_{2}, \ldots, \mathbf{x}_{m} \}_j\) be the set of genuine feature vectors of user $j$, with $j = 1, \ldots, N$. Next, a writer-related centroid $\boldsymbol{\mu}$ (local centroid) is computed as defined in Equation~\ref{eq:eq_local_centroid}.

\begin{equation}
\label{eq:eq_local_centroid}
 \boldsymbol{\mu}_j = \frac{1}{m} \sum_{i=1}^{m} \mathbf{x}_i, 
 \quad \mathbf{x}_i \in \mathcal{X}_j, \quad 
 \boldsymbol{\mu}_j,\, \mathbf{x}_i \in \mathbb{R}^d
\end{equation}

Subsequently, $\boldsymbol{\mu}_j$ is employed to obtain the closest negative samples represented by the prototypical signature vectors. Formally, let $\mathcal{C} = \{\mathbf{c}_1, \mathbf{c}_2, \dots, \mathbf{c}_k\}$ be the set of all prototypical signatures produced by the data-driven summarization process, the distance-based selection of negative samples is computed according to Equation~\ref{eq:eq_prototype_selection}. This formulation selects the $h$ prototypical signature vectors in $\mathcal{C}$ that are closest to the local centroid $\boldsymbol{\mu}_j$ of user $j$ under the Euclidean norm. 

\begin{equation}
\label{eq:eq_prototype_selection}
\mathcal{P}_j = \operatorname{sort}_{\text{asc}, h} 
\left[ 
\left( \|\mathbf{c}_i - \boldsymbol{\mu}_j\|_2 \right)_{i=1}^k 
\right], \quad |\mathcal{P}_j| = h, \quad \mathbf{c}_i \in \mathcal{C}
\end{equation}

where $\mathcal{P}_j$ denote the set of selected prototypical signatures for user $j$. These selected samples serve as negative examples and are combined with the genuine samples $\mathcal{X}_j$ to form the training set $\mathcal{T}_j$. 

\begin{algorithm}
\caption{Training Data Generation}
\label{algo:data_generation}
\begin{algorithmic}[1]
\REQUIRE $N$ users, feature vectors $\mathcal{X}$, all prototypical signature vectors $\mathcal{C}$, number of negative samples to be selected $h$
\STATE Initialize final training set: $\mathcal{T} \gets \emptyset$
\FOR{each user $j \in \{1, \ldots, N\}$}
    \STATE Retrieve feature vectors: $\mathcal{X}_j$
    \STATE Compute local centroid: $\mu_j$ (Equation~\ref{eq:eq_local_centroid})
    \STATE Select $h$ closest prototypical signature to $\mu_j$ (Equation~\ref{eq:eq_prototype_selection})
    
    \STATE Label vectors in $\mathcal{X}_j$ as positive 
    \STATE Label vectors in $\mathcal{P}_j$ as negative
    
    \STATE Form training set:  $\mathcal{T}_j \gets \mathcal{X}_j \cup \mathcal{P}_j$
    \STATE Update final training set: $\mathcal{T} \gets \mathcal{T}  \cup \mathcal{T}_j$
\ENDFOR
\STATE \textbf{return} $\mathcal{T}$
\end{algorithmic}
\end{algorithm}
 
\subsection{WI-Classifier Training}

The writer-independent setting requires transforming the multi-class signature verification problem into a binary classification task. To this end, we employ the Dichotomy Transformation (DT) \cite{Souza2019_CharacterizationOfHSinDissimilarityRepresentationSpace}, which computes dissimilarity vectors between pairs of feature vectors.
Suppose two feature vectors $\mathbf{x}_R$ and $\mathbf{x}_C$,  with $\mathbf{x}_R = \{f_{i}^{R}\}_{i=1}^{d}$ and $\mathbf{x}_C = \{f_{i}^{C}\}_{i=1}^{d}$, where $d$ is the number of features $f$. The dissimilarity vector between $\mathbf{x}_R$ and $\mathbf{x}_C$ is given by $\mathbf{\dot{x}}_{RC}=DT(\mathbf{x}_R,\mathbf{x}_C) = \{|f_{i}^{R}- f_{i}^{C}|\}_{i=1}^{d}$, where $\left| \: \cdot \: \right|$ represents the absolute value of the difference. The vector $\mathbf{\dot{x}}_{RC}$ has the same dimensionality as $\mathbf{x}_R$ and $\mathbf{x}_C$. If $\mathbf{\dot{x}}_{RC}$ is obtained from signatures of the same user, it is labeled as \textit{positive}. Otherwise, it is labeled as \textit{negative}.  

In this work,  given the set of genuine feature vectors \( \mathcal{X}_j = \{\mathbf{x}_1, \mathbf{x}_2, \ldots, \mathbf{x}_m\}_j \), and the set of selected prototypical signatures \( \mathcal{P}_j = \{\mathbf{p}_1, \mathbf{p}_2, \ldots, \mathbf{p}_h\}_j \) for user \( j \), we construct a dissimilarity-based training set \( \dot{\mathcal{Y}} \) by applying the dichotomy transformation \( \text{DT} \) to all pairs \((\mathbf{x}_i, \mathbf{x}_t) \) of genuine feature vectors, and all pairs \( (\mathbf{x}_i, \mathbf{p}_t) \) of genuine and prototypical signature vectors, where $\mathbf{x}_i$ and $\mathbf{x}_t \in \mathcal{X}_j$ and  $\mathbf{p}_t \in \mathcal{P}_j$. Formally, considering $N$ users contributing with $m$ signatures each, the resulting set of dissimilarities is defined by Equations~\ref{eq:eq_diss_set} and \ref{eq:eq_diss_set_user}.

\begin{equation}
\label{eq:eq_diss_set}
   \dot{\mathcal{Y}} = \bigcup_{j=1}^{N} \left\{  \dot{\mathcal{Y}}_j\right\}
\end{equation}

\begin{equation}
\label{eq:eq_diss_set_user}
   \dot{\mathcal{Y}}_j = \bigcup_{i=1}^{m} \bigcup_{t=i+1}^{m-1} \left\{ \text{DT}(\mathbf{x}_i, \mathbf{x}_t) \right\} \quad \cup \quad  \bigcup_{i=1}^{m} \bigcup_{t=1}^{h} \left\{ \text{DT}(\mathbf{x}_i, \mathbf{p}_t) \right\}
\end{equation}

Where  \( \dot{\mathcal{Y}}_j \) is formed by two terms. The first term computes all pairwise dissimilarities among genuine signatures of user $j$. Since these pairs come from the same writer, they represent the \textit{positive} class. This operation generates $\binom{m}{2}$ positive dissimilarity vectors per user. The second term computes dissimilarities between each genuine signature $\mathbf{x}_i \in \mathcal{X}_j$ and the $h$ prototypical signatures $\mathbf{p}_t \in \mathcal{P}_j$. Since prototypes serve as forgery signatures, these pairs represent the \textit{negative} class. This produces $m \times h$ negative dissimilarity vectors per user.

Therefore, $\dot{\mathcal{Y}}_j$ contains both the \textit{within-user} similarities (positives) and the \textit{between-user} comparisons with prototypical signatures (negatives). 
By combining all users’ sets $\dot{\mathcal{Y}}_j$, we obtain the global dissimilarity set $\dot{\mathcal{Y}}$ in Eq.~\ref{eq:eq_diss_set}, which is then used to train the writer-independent classifier.

\section{Experimental Setup}
\label{sec:ex_protocl}

\noindent
\textbf{Datasets and segmentation.} Experiments are performed on the datasets described in Table~\ref{tab:datasets}. Each dataset is partitioned into a development set $\mathcal{D}$ for classifier training, and an exploitation set $\mathcal{E}$ for testing.

\begin{table}[ht!]
\caption{Datasets used in this work and their user segmentation. G denotes genuine signatures. SK denotes skilled forgery signatures.}
\label{tab:datasets}
\centering
\resizebox{\textwidth}{!}{\begin{tabular}{lccc ccc ccc}
\hline
\multirow{2}{*}{Dataset} & \multicolumn{2}{c}{Signatures} & & \multicolumn{2}{c}{Development set $\mathcal{D}$} & & \multicolumn{2}{c}{Exploitation set $\mathcal{E}$} \\
\cline{2-3} \cline{5-6} \cline{8-9}
 & Users & G / SK & & Users & ID range & & Users & ID range \\ \cline{1-1} \cline{2-3} \cline{5-6} \cline{8-9}
CEDAR \cite{Kalera2004_DB_CEDAR}         & 55    & 24 / 24 & & 27  & 1--27    & & 28   & 28--55 \\
MCYT-75 \cite{OrtegaGarcia2003_DB_MCYT}       & 75    & 15 / 15 & & 37  & 1--37    & & 38   & 38--75 \\
GPDS Synthetic \cite{Ferrer2015_DB_GPDSsynthetic} & 10000 & 24 / 30 & & 581 & 301--881 & & 300  & 1--300 \\
\hline
\end{tabular}
}
\end{table}

\noindent \textbf{Data generation.} To evaluate the proposed system, classifiers are trained and tested on dissimilarity vectors derived from the dichotomy transformation. We follow the signature segmentation employed in \cite{Talles2023_AmultitaskApproach4ContrastiveLearning} for creating the dissimilarity sets as defined in Table~\ref{tab:wi_dev_exp_sets}.

\begin{table}[ht!]
\centering
\caption{Signatures segmentation for WI approach.}
\label{tab:wi_dev_exp_sets}
\begin{subtable}{\textwidth}
\centering
\caption{Development set ($\mathcal{D}$).}
\label{tab:wi_development_set}
\begin{tabular}{
  >{\raggedright\arraybackslash}p{2cm}
  >{\centering\arraybackslash}p{0.5cm}
  >{\raggedright\arraybackslash}p{5cm}
  >{\raggedright\arraybackslash}p{4cm}
}
\toprule
 \textbf{Method} & & \multicolumn{1}{c}{\textbf{Negative class}} & \textbf{Positive class} \\
  \textbf{} & & \multicolumn{1}{c}{\textit{Distance between:}} & \textit{Pairwise distance of:} \\
\midrule

 Standard && 6 random forgeries and 11 genuine signatures for each user  & \multirow{4}{=}{12 genuine signatures for each user}  \\ Summarization && 6 prototypical signatures and 11 genuine signatures for each user & \\ \bottomrule
\end{tabular}
\end{subtable}


\begin{subtable}{\textwidth}
\centering
\caption{Exploitation set ($\mathcal{E}$).}
\label{tab:wi_exploitation_set}
\begin{tabular}{
  >{\raggedright\arraybackslash}p{0.15\textwidth}
>{\raggedright\arraybackslash}p{0.3\textwidth}
  >{\raggedright\arraybackslash}p{0.5\textwidth}
}
\toprule
\textbf{Data}  
& \textbf{Reference set} 
& \textbf{Claimed set}  \\
\midrule

CEDAR
& $r \in \{1, 2, 3, 5, 10, 12\}$ 
& 10 genuine, 10 random, 10 skilled\\

GPDS-S
& $r \in \{1, 2, 3, 5, 10, 12\}$ 
& 10 genuine, 10 random, 10 skilled\\

MCYT
& $r \in \{1, 2, 3, 5, 10\}$ 
& 5 genuine, 5 random, 5 skilled\\

\bottomrule
\end{tabular}
\end{subtable}
\end{table}

For the signature segmentation in the development set $\mathcal{D}$ (Table~\ref{tab:wi_development_set}), we employ two configurations: the standard (baseline), which utilizes genuine signatures from other users (random forgeries) to create negative dissimilarities, and the proposed method, which utilizes prototypical signatures instead. To produce a balanced dataset,  twelve genuine signatures are randomly selected to make the positive samples, while eleven signatures are used against six random forgeries or prototypical signatures to form the negative class.

For the signature segmentation in the exploitation set (Table~\ref{tab:wi_exploitation_set}), we use varying numbers of reference signatures, from 1 to 12, randomly chosen for each user. These references are then compared against genuine signatures (yielding positive test samples) and against skilled forgery signatures (yielding negative test samples).

\noindent \textbf{Preprocessing.} We follow the steps described in \cite{Talles2023_AmultitaskApproach4ContrastiveLearning}. Specifically, signature images are initially centered on a canvas of size equal to that of the largest sample. The background is removed using Otsu’s algorithm, setting it to white and the foreground to grayscale. The image is then inverted, resized to 170×242 pixels, and center-cropped to 150×220 pixels. 

\noindent \textbf{Feature Extraction.} We utilize SigNet Synthetic (\textit{SigNet-S}) \cite{Talles2023_AmultitaskApproach4ContrastiveLearning}, 
a deep convolutional neural network specifically developed to learn discriminative characteristics of individual signatures. The model was trained on the GPDS Synthetic dataset~\cite{Ferrer2015_DB_GPDSsynthetic} employing user IDs 5001–7000, ensuring a subset disjoint from the data used in our experiments. Feature extraction is performed by forwarding each signature image through the network, yielding a 2048-dimensional feature vector.

\noindent \textbf{Classifiers.} In this work, we employ two classification approaches for the verification step: (i) an  SVM with a radial basis function kernel, hereafter referred to as SVM-RBF; and (ii) a linear SVM trained via the primal formulation using stochastic gradient descent, hereafter referred to as SVM-Linear (SGD). Both classifiers are implemented using the \texttt{scikit-learn} library \cite{sklearn_api}. The SGD classifier uses a hinge loss function with an $\alpha$ parameter, which controls the regularization strength, set to $0.1$, a convergence tolerance of $0.001$, and a maximum of $2000$ iterations, following \cite{deMoura2024_shsv}. The SVM-RBF configuration, based on \cite{Talles2023_AmultitaskApproach4ContrastiveLearning}, uses a regularization parameter $C = 1.0$ and an RBF kernel coefficient $\gamma = 2^{-11}$.

\noindent
\textbf{Defining the summarization hyperparameter $k$.} The value of $k$ was determined using cross-validation on the development set of each dataset. The optimal values were found to be 150, 10, and 50 for SVM-RBF, and 100, 10, and 100 for SVM-Linear (SGD), corresponding to GPDS-S, CEDAR, and MCYT, respectively. We also conducted a sensitivity analysis of the hyperparameter $k$, as it is an essential component of our system. Results demonstrate robustness with respect to the choice of $k$, with the model delivering competitive and stable performance across a broad range of $k$ values. A detailed description of the validation protocol and sensitivity analysis is provided in \href{wi_proto_hsv_supp.pdf#supp_wd}{Sections 3 and 4} of the supplementary material.

\noindent
\textbf{Performance evaluation.} Performance is evaluated using the Equal Error Rate (EER), defined as the point where the False Rejection Rate (FRR) is equal to the False Acceptance Rate (FAR) \cite{Bolle2000_eer}. Each experiment is repeated five times. In every repetition, a new random subset of signature samples is selected, and prototypical signatures are chosen based on distance measures. The EER is calculated using both global and user-specific thresholds based on the distance scores to the decision hyperplane. The final reported performance is the mean and standard deviation of the EER across all independent repetitions.

\noindent
\textbf{Model Complexity Evaluation.} To evaluate the system’s computational complexity and scalability, we assess multiple aspects, including runtime performance, memory usage, computational cost, and arithmetic intensity, using the following metrics:

\begin{itemize}[topsep=0pt,
    label=\textbullet]
     \item \textbf{Training and testing time (in seconds):} Measures the total time required for model training and prediction.
    \item \textbf{Model size (in megabytes):} Evaluates the trained model's total storage. 
     \item \textbf{Number of support vectors (SV):} Indicates model complexity, since more support vectors generally increase memory and prediction cost.
   
    \item \textbf{Training and testing FLOPs (floating-point operations)}: Quantifies the computational cost for training and prediction.
\end{itemize}
 
To compute FLOPs, we utilized the PAPI\footnote{https://github.com/icl-utk-edu/papi} library \cite{Jagode2025_papi} through the Python wrapper PYPAPI\footnote{https://github.com/flozz/pypapi}, which enables access to hardware performance counters. FLOP counts were measured separately for training and testing, isolated from any unrelated background processes. The complete machine configuration used to run the experiments can be found in \href{wi_proto_hsv_supp.pdf#supp_arch}{Section 5} of the supplementary material.

All complexity measurements were averaged over five runs to mitigate variability due to system load or caching effects.

\section{Results}

This section presents experimental evidence supporting our main claims: (1) the proposed summarization-based training method performs on par with or surpasses the standard random sampling approach; (2) it substantially reduces the development set size and computational demands, thereby enabling scalable deployment, and (3) it operates independently of the underlying backbone architecture, allowing seamless integration with diverse feature extractors.

\vspace{1em}
\noindent
\textbf{Performance with Prototypical Signatures.} The results for performance verification are presented in Figure~\ref{fig:wi_eer_best_k_signets_skilled}\footnote{Tabular version of results are provided in Section~6 of the supplementary material.}
which shows the equal error rate as a function of the number of reference signatures used. As demonstrated, our proposed method for generating dissimilarities outperforms the standard approach in most cases when using global thresholds, regardless of the number of reference samples.

\begin{figure}[ht!]
    \centering
    \includegraphics[
    trim=0 0 0 30, clip,
    width=\linewidth]{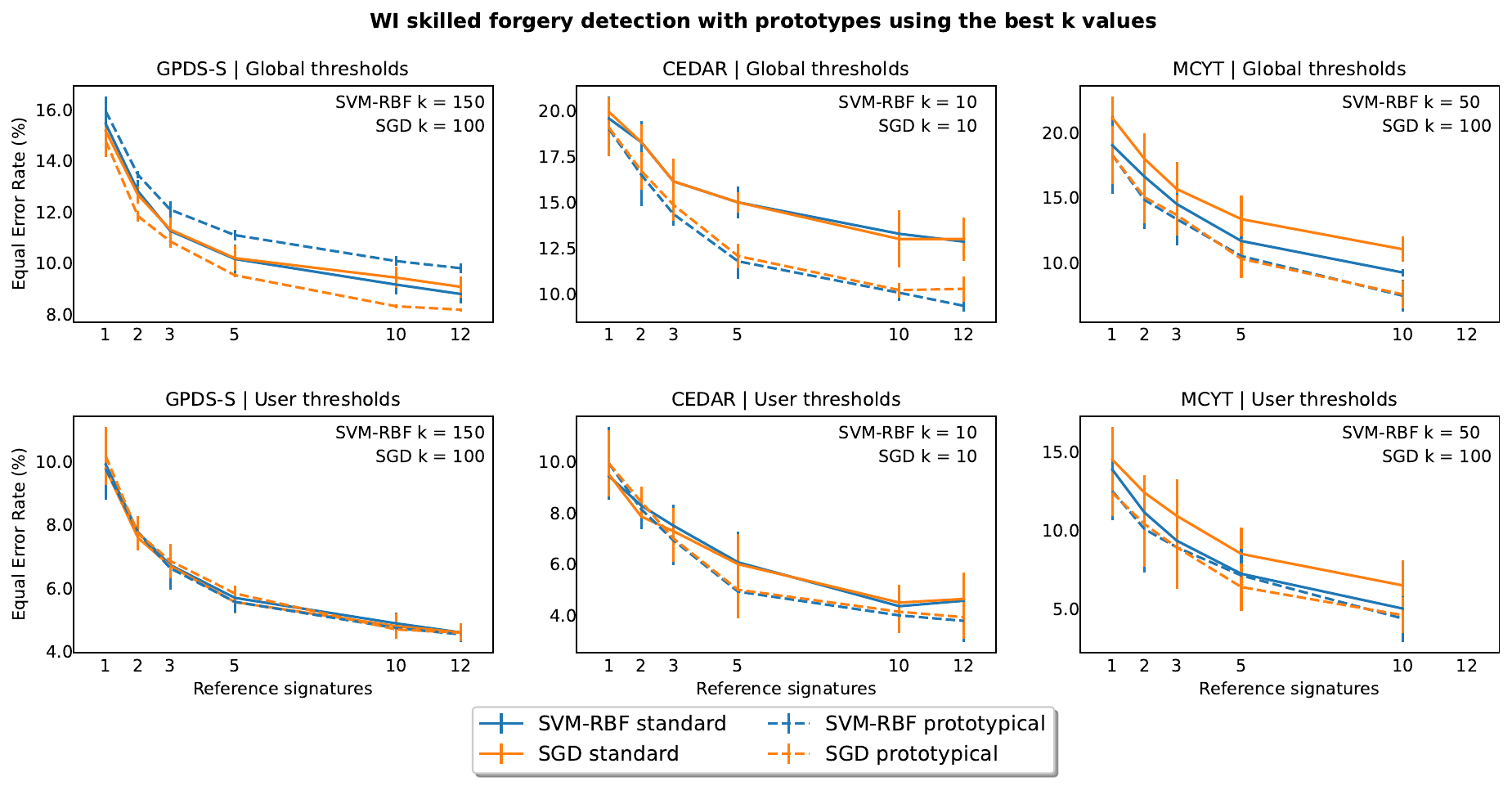}
     \caption{WI average EER for skilled forgery detection across different reference signatures using the best $k$ values for SVM-RBF and SVM-Linear (SGD).}
    \label{fig:wi_eer_best_k_signets_skilled}
\end{figure}

\begin{figure}[h!]

    \centering
    \includegraphics[
width=\linewidth]{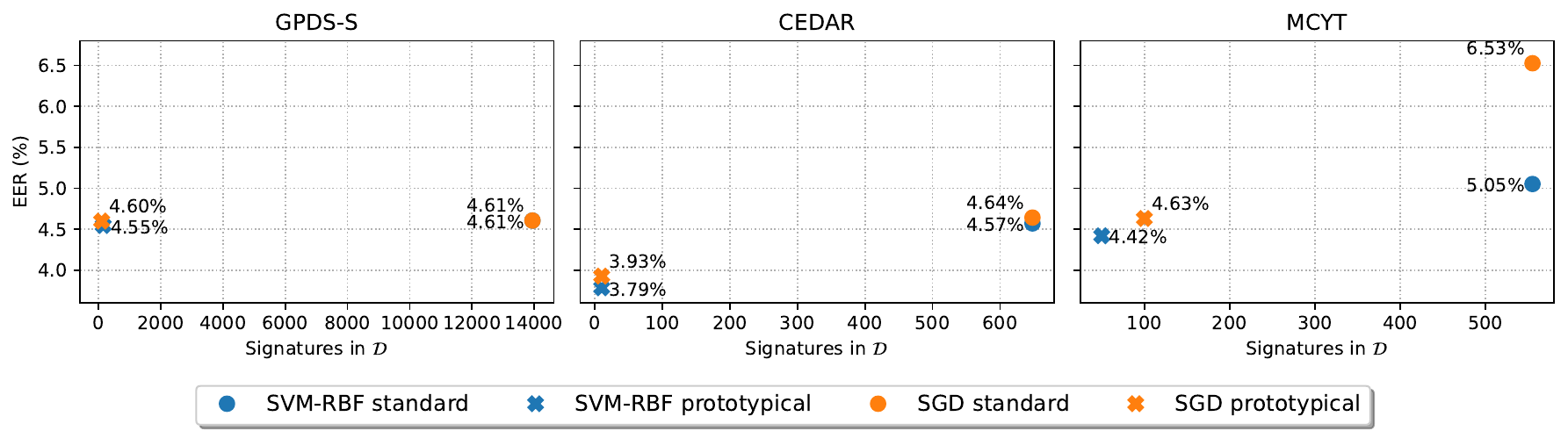}
     \caption{WI average EER vs. the total number of signatures in the development set for the standard approach (circles) and the prototypical method (crosses). Results correspond to skilled forgery detection using the maximum number of reference signatures and user-specific thresholds. For the prototypical method, results are reported with the best $k$ value selected on each dataset for SVM-RBF and SVM-Linear (SGD).}
    \label{fig:wi_samples}
\end{figure}

Furthermore, our proposed method yields comparable results to the standard approach, even after reducing the entire development set. This can be observed in Figure~\ref{fig:wi_samples}, which presents the EER vs. the total number of signatures in the $\mathcal{D}$ set for the standard approach (circles) and for the prototypical method (crosses).  

For the GPDS-S dataset, for example, which comprises 13,944 signatures (24 $\times$ 581), the summarization technique downsizes it to only 100 samples for SVM-Linear (SGD) and 150 for SVM-RBF, which represents a reduction of more than  98.9\%. For CEDAR and MCYT, the reduction is, respectively, of 98.5\% and 91\% for SVM-RBF; and of  98.5\% and 82\% for SVM-Linear (SGD). Thus, the proposed method matches or exceeds the standard approach while significantly reducing the set size from which negative samples are selected.

\vspace{1em}
\noindent
\textbf{Prototypical Signatures with Linear Classifier Significantly Reduce Computational Complexity.} As previously demonstrated, the prototypical approach also performs very well with a linear SVM trained using SGD for offline handwritten signature verification, often matching or surpassing the performance of the widely adopted SVM-RBF. \emph{But what are the practical advantages of this substitution, and why does it matter?} The answer lies in computational complexity.

\begin{figure}[h!]
    \centering
    \includegraphics[width=\linewidth]{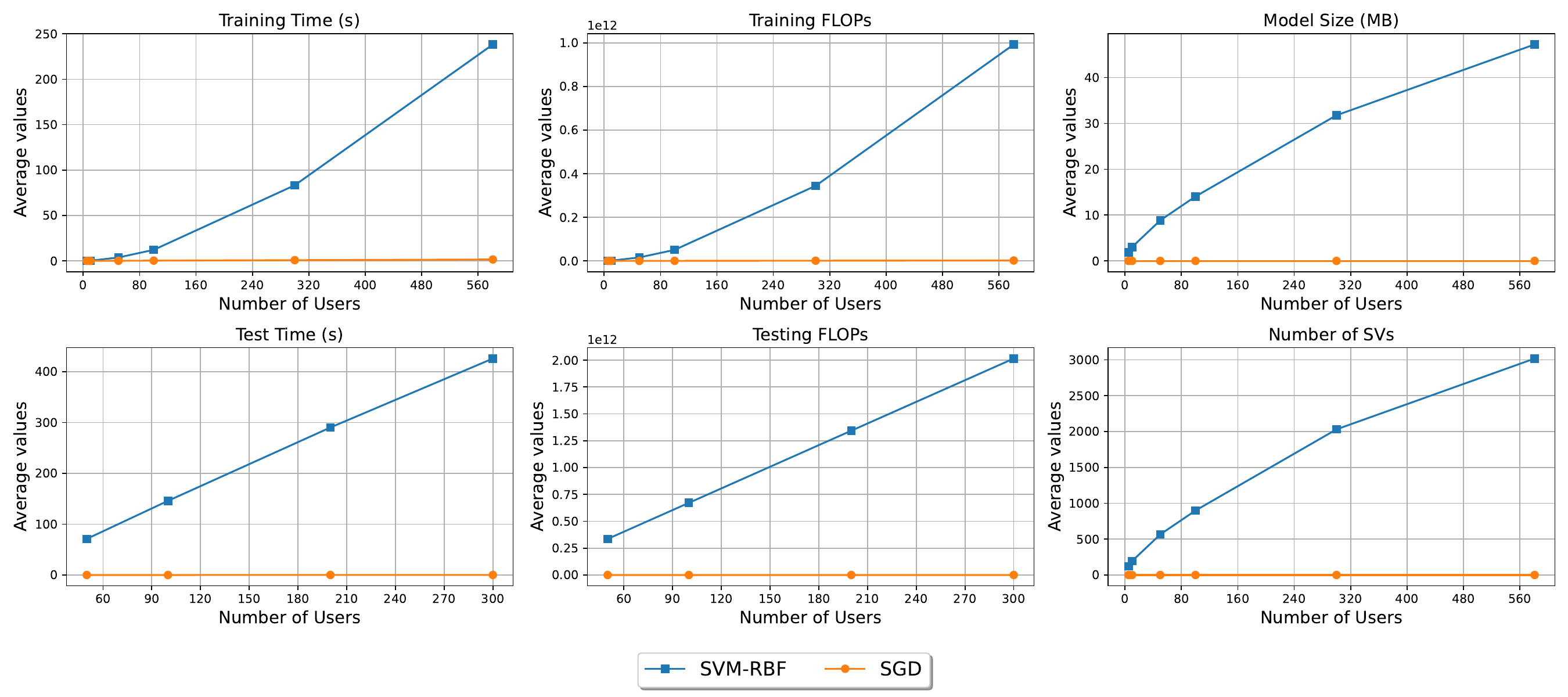}
     \caption{Computational cost of SVM-RBF and SVM-Linear (SGD) in WI signature verification as the number of users increases in GPDS-S dataset.}
\label{fig:wi_running_metrics}
\end{figure}

Figure~\ref{fig:wi_running_metrics} presents a comparison between an SVM with a radial basis function kernel and a linear SVM optimized with SGD under identical settings, illustrating the evolution of different running time metrics as the number of users increases. Additionally, Table~\ref{tab:model_comparison} summarizes the values obtained when using the largest number of users for training and testing.

\begin{table}[ht!]
    \centering
    \caption{WI computational cost comparison between SVM-RBF and SVM-Linear (SGD) with the whole GPDS-S $\mathcal{D}$ and $\mathcal{E}$ sets.}
    \label{tab:model_comparison}
    \begin{tabular}{@{}lcc@{}}
        \toprule
        \textbf{Metric} & \textbf{SVM-RBF} & \textbf{SVM-Linear (SGD)} \\ \midrule
        \textbf{Training Time (s)} & 238.31 & 1.59 \\
        \textbf{Testing Time (s)} & 425.95 & 0.11 \\
        \textbf{Model Size (KB)} & 47,205.95 & 8.91 \\
        \textbf{Number of SVs} & 3017 & 0 \\

        \textbf{Training FLOPs} & $9.93 \times 10^{11}$ & $1.88 \times 10^{9}$ \\
\textbf{Testing FLOPs}  & $2.02 \times 10^{12}$ & $1.10 \times 10^{5}$ \\

        \bottomrule
    \end{tabular}
\end{table} 
As shown, the differences are substantial. SVM-RBF training required $9.93 \times 10^{11}$ FLOPs on average, with training time increasing substantially with dataset size. The number of support vectors grows rapidly, increasing model complexity and resource demands. For non-linear kernels, the SVM solver\footnote{The authors in \cite{sklearn_api} based SVM's implementation on LibSVM\cite{chang2011_libsvm} which employs a version of the Sequential Minimal Optimization algorithm presented in \cite{Chen2006_SMO_SVM}.} used has computational complexity between $\mathcal{O}(n^2 \cdot d)$ and $\mathcal{O}(n^3 \cdot d)$, where $n$ is the number of samples and $d$ the number of features \cite{sklearn_api}, making it computationally expensive and challenging to scale.

In contrast, the SVM-Linear (SGD) model requires far less computation. For inference, while SVM-RBF evaluates 3017 support vectors of 2048 dimensions each (about 6.17 million operations), the SGD-based model performs a single 2049-dimensional dot product (2048 weights plus a bias term), offering an estimated 3000× speedup. As shown in Table~\ref{tab:model_comparison}, SVM-Linear (SGD) training time averaged only 1.59 seconds, and testing time 0.11 seconds when the whole $\mathcal{D}$ and $\mathcal{E}$ sets are used.

The model size also highlights the scalability advantage of SVM-Linear (SGD). As the number of users grows, the computational burden and model size of SVM-RBF increase steeply, as seen in Figure~\ref{fig:wi_running_metrics}. This makes the method inherently non-scalable. The growth in support vectors directly inflates both training cost and storage, with model size reaching 47,205.95 KB in our largest setting. In contrast, the linear SVM (SGD) remains compact (8.91 KB) and unaffected by dataset size, making it far more scalable for large-scale deployments.

While current studies are typically confined to lab-scale datasets of hundreds to thousands of writers \cite{Singla2025_HSV_survey}, our approach enables scalable implementations for real-world applications involving millions of users, such as vote-by-mail elections where officials must verify voters' handwritten signed ballots before being counted \cite{Janover2020_caseStudyCalifornia}, and the volume can far exceed millions\footnote{In the 2024 U.S. election, states reported that 46,846,449 voters cast mail ballots that were counted \cite{EAVS_2024}.}, which is beyond the limitations of current works.

\noindent
\textbf{A Backbone-Agnostic Method.} Finding efficient representations of signature images is widely explored in the literature \cite{LI2024_TransOSV,REN2023_2C2S_twoChannelAndTwoStreamTransformerBased,Tsourounis2022_singleSignature,Tsourounis2025_FKD_WithoutSignatures,Zhang2024_RF_HSV_Disentangling}, with significant advances in recent years driven by deep learning methods \cite{Singla2025_HSV_survey}. To demonstrate that our proposed approach operates independently of the backbone employed, we evaluated several feature extractors under a WI configuration. 

We employed models with different architectures from two recent works, with publicly available and reproducible code: \cite{Tsourounis2025_FKD_WithoutSignatures} and \cite{Talles2024_ContinualLearning_csigver}. In \cite{Tsourounis2025_FKD_WithoutSignatures}, Feature Knowledge Distillation (FKD) transfers knowledge from a teacher (SigNet) to a student (ResNet18) without using signature images. The study evaluated several strategies, geometric distillation (GEOM) for local alignment and three global objectives: temperature-scaled cross-entropy (T-CE), Barlow Twins (BT), and the proposed Barlow Colleagues (BC). As combining local (GEOM) and global (classification-based) KD proved most effective, we adopted three backbones: GEOM \& TCE, GEOM \& BT, and GEOM \& BC. In \cite{Talles2024_ContinualLearning_csigver}, continual learning with knowledge distillation was applied to improve real-signature representations by generating synthetic examples that complement real data. Knowledge from a teacher (SigNet) is distilled into student models using a joint Kullback–Leibler and cross-entropy loss. We used the resulting architectures: Continual SigNet (AlexNet-based), Continual ResNet152, and Continual Vision Transformer (ViT).
 
We followed the experimental protocol described in Section~\ref{sec:ex_protocl}, and repeated the experiment ten times as in the original work to enable statistical analysis. The evaluation was conducted on the same datasets used in \cite{Tsourounis2025_FKD_WithoutSignatures} (CEDAR and MCYT) and in \cite{Talles2024_ContinualLearning_csigver} (GPDS-S, CEDAR, and MCYT). Table~\ref{tab:comparison_wi_bb_user} shows the results for user-specific thresholds. The complete validation process and results for the global threshold are provided in \href{wi_proto_hsv_supp.pdf#supp_bb}{Section~7} of the supplementary material.

\begin{table}[htbp]
\caption{Performance across different backbones in a WI setting. All experiments employed an SVM-RBF with the standard negative sampling and a linear SVM optimized with SGD using prototypical signatures. Reported results show $EER$ with user thresholds. 
The $p$-values correspond to a paired \textit{t}-test if normality (Shapiro–Wilk\tablefootnote{\label{fn:shapiro}Shapiro–Wilk test results are provided in \href{WI_HSV_protoypes_supp.pdf\#supp_bb}{Section~7} of the supplementary material.}) was satisfied, or to a Wilcoxon signed-rank test otherwise. ROPE is defined as $[-0.015,\,0.015]$. Results demonstrating statistical equivalence are highlighted.}
\label{tab:comparison_wi_bb_user}
\centering
\resizebox{\textwidth}{!}{\begin{tabular}{l l l c c c r}
\toprule
\multicolumn{1}{c}{\multirow{2}{*}{Dataset}} & \multicolumn{1}{c}{\multirow{2}{*}{Model}} & \multicolumn{2}{c}{$EER_{user}$ (\%)} & \multicolumn{1}{c}{\multirow{2}{*}{p-value}} & \multicolumn{1}{c}{\multirow{2}{*}{\% in ROPE}} & \multicolumn{1}{c}{\multirow{2}{*}{95\% HDI}}  \\ \cline{3-4}
&  & Standard & Prototypical &  &  & \\
\midrule
 \multirow[t]{6}{*}{CEDAR} & ResNet18 CL + KD: GEOM with BC & 2.39 $\pm$ 0.48 & 1.93 $\pm$ 0.43 & \textcolor{black}{0.04} & \textcolor{black}{\textbf{100.0}} & \boldmath{$-0.009\; \text{--}\;0.000$}  \\
  & ResNet18 CL + KD: GEOM with BT & 1.29 $\pm$ 0.46 & 1.29 $\pm$ 0.83 & \textcolor{black}{1.00} & \textcolor{black}{\textbf{99.3}} & \boldmath{$-0.009\; \text{--}\;0.009$}  \\
  & ResNet18 CL + KD: GEOM with TCE & 1.43 $\pm$ 0.53 & 1.93 $\pm$ 0.74 & \textcolor{black}{0.20} & \textcolor{black}{\textbf{98.5}} & \boldmath{$-0.004\; \text{--}\;0.014$}  \\
  & Continual ResNet152 & 2.64 $\pm$ 0.68 & 2.93 $\pm$ 0.76 & \textcolor{black}{0.16} & \textcolor{black}{\textbf{99.0}} & \boldmath{$-0.007\; \text{--}\;0.012$}  \\
  & Continual SigNet & 2.61 $\pm$ 0.42 & 2.39 $\pm$ 0.60 & \textcolor{black}{0.53} & \textcolor{black}{\textbf{99.7}} & \boldmath{$-0.010\; \text{--}\;0.006$}  \\
  & Continual ViT & 3.14 $\pm$ 0.75 & 3.18 $\pm$ 0.44 & \textcolor{black}{0.91} & \textcolor{black}{\textbf{99.9}} & \boldmath{$-0.007\; \text{--}\;0.008$}  \\
\addlinespace[0.5em] \multirow[t]{3}{*}{GPDS-S} & Continual ResNet152 & 4.02 $\pm$ 0.32 & 4.21 $\pm$ 0.28 & \textcolor{black}{0.09} & \textcolor{black}{\textbf{100.0}} & \boldmath{$-0.001\; \text{--}\;0.004$}  \\
  & Continual SigNet & 4.18 $\pm$ 0.31 & 4.55 $\pm$ 0.20 & \textcolor{black}{0.02} & \textcolor{black}{\textbf{100.0}} & \boldmath{$0.001\; \text{--}\;0.007$}  \\
  & Continual ViT & 5.69 $\pm$ 0.34 & 7.65 $\pm$ 0.44 & \textcolor{black}{0.00} & \textcolor{black}{3.0} & $0.015\; \text{--}\;0.025$  \\
\addlinespace[0.5em] \multirow[t]{6}{*}{MCYT} & ResNet18 CL + KD: GEOM with BC & 4.16 $\pm$ 0.89 & 5.42 $\pm$ 1.27 & \textcolor{black}{0.02} & \textcolor{black}{66.9} & $0.002\; \text{--}\;0.024$  \\
  & ResNet18 CL + KD: GEOM with BT & 4.32 $\pm$ 1.45 & 3.95 $\pm$ 1.34 & \textcolor{black}{0.59} & \textbf{91.6} & \boldmath{$-0.015\; \text{--}\;0.013$}  \\
  & ResNet18 CL + KD: GEOM with TCE & 4.26 $\pm$ 0.83 & 5.68 $\pm$ 1.28 & \textcolor{black}{0.00} & \textcolor{black}{58.4} & $0.005\; \text{--}\;0.023$  \\
  & Continual ResNet152 & 3.53 $\pm$ 1.45 & 4.84 $\pm$ 0.77 & \textcolor{black}{0.06} & \textcolor{black}{61.0} & $-0.001\; \text{--}\;0.027$  \\
  & Continual SigNet & 2.68 $\pm$ 1.12 & 3.11 $\pm$ 0.95 & \textcolor{black}{0.47} & \textbf{93.9} & \boldmath{$-0.009\; \text{--}\;0.014$}  \\
  & Continual ViT & 6.32 $\pm$ 1.37 & 9.63 $\pm$ 2.05 & \textcolor{black}{0.00} & \textcolor{black}{1.5} & $0.017\; \text{--}\;0.049$  \\
\bottomrule
\end{tabular}
}
\end{table} 
We assessed statistical equivalence between configurations (with and without prototypical signatures) across datasets and models. Residual normality was tested using the Shapiro–Wilk test\footref{fn:shapiro} \cite{Shapiro1965_normality}, followed by a paired t-test \cite{Student1908} or Wilcoxon signed-rank test \cite{Wilcoxon1945}, as appropriate; results are shown in the “p-value” column. To complement the frequentist analysis, we performed a Bayesian inference using the highest density interval (HDI) and a Region of Practical Equivalence (ROPE) of $[-0.015, 0.015]$, with a 95\% credible interval \cite{Kruschke2018_TheBayesianNewStatistics}.

As can be observed, employing prototypical signatures with a linear SVM achieves performance comparable to that of an SVM with an RBF kernel across most datasets and backbones. The effectiveness of the method holds in a backbone-agnostic manner, with architectures such as SigNet, ResNet, and ViT all achieving statistically equivalent performance under several configurations.

\section{Conclusion}
\label{sec:conclusion}

This work introduced a data generation strategy for handwritten signature verification that summarizes development data into prototypical signatures, which are then used as negative samples. Unlike random forgeries drawn arbitrarily from other users, the proposed method selects prototypical signatures based on their distance to each user’s genuine signatures. Experiments on performance, scalability, and backbone integration show that the approach produces more informative negatives, matches or exceeds traditional methods, and greatly improves efficiency when combined with linear SVMs trained via SGD. Moreover, it operates in a backbone-agnostic manner, ensuring robust results across feature extractors. Limitations of the method include reliance on clustering quality, potential bias in the common-pattern representation, and the need for sufficient sample size. Future work will explore alternative clustering methods, improved hyperparameter tuning, and adaptive learning of prototypical signatures.

\bibliographystyle{plain}
\bibliography{references}



\section*{Supplementary material}

# A Prototypical Signature Approach for Writer-Independent Offline Signature Verification – Supplementary Material –

Kecia G. de Moura, Robert Sabourin, and Rafael M. O. Cruz

École de technologie supérieure - Université du Québec, Montreal, Québec, Canada
kecia.gomes-de-moura.1@ens.etsmtl.ca
{rafael.menelau-cruz,robert.sabourin}@etsmtl.ca

## 1 Overview

This supplementary document provides additional material that complements the main manuscript. Section 2 lists the symbols and notation used throughout the study for clarity and consistency. Section 3 presents the procedure employed to determine the optimal number of prototypical signatures ($k$), and Section 4 analyzes the system's sensitivity to different $k$ values. Section 5 describes the hardware configuration used for computational complexity evaluation. Section 6 presents the tabular results of the system's performance using prototypical signatures. Finally, Section 7 reports the results obtained when employing different backbone architectures using a global threshold.

## 2 Symbols and Notation

To facilitate clarity and readability, Table 1 lists the symbols and notation used throughout the study.

Table 1: Summary of notation and symbols used in the paper.

| Symbol | Description | Symbol | Description |
|---|---|---|---|
| $S$ | Signature image | $\phi(\cdot)$ | Feature extractor |
| $DT(\cdot,\cdot)$ | Dichotomy transformation | $\theta$ | Classifier |
| $\mathbf{x}$ | Feature vector $\mathbf{x} = \phi(S)$ | $\dot{\mathbf{x}}$ | Dissimilarity vector $\dot{\mathbf{x}} = DT(\mathbf{x_1}, \mathbf{x_2})$ |
| $y$ | Dissimilarity label | $G$ | Genuine signature |
| $RF$ | Random forgery signature | $SK$ | Skilled forgery signature |
| $R$ | Reference signature | $C$ | Claimed signature |
| $\mathcal{S}$ | Set of signatures | $\mathcal{X}$ | Set of feature vectors |
| $\dot{\mathcal{Y}}$ | Set of dissimilarity vectors | $\mathcal{D}$ | Development set |
| $\mathcal{E}$ | Exploitation set | $\mathcal{C}$ | Set of all prototypical signatures |
| $\mathcal{P}$ | Set of selected prototypical signatures | $\mathcal{K}(\cdot,\cdot)$ | Data-driven summarization |
| $\mathbf{c}$ | Vector $\in \mathcal{C}$ | $\boldsymbol{\mu}$ | Local centroid vector |
| $\mathcal{T}$ | Final training data | $\mathbf{p}$ | Vector $\in \mathcal{P}$ |

## 3 Finding the Best $k$ Value

The number of partitions $k$ is an essential hyperparameter to be defined in order to properly adjust the proposed system. To do so, a 10-fold cross-validation is

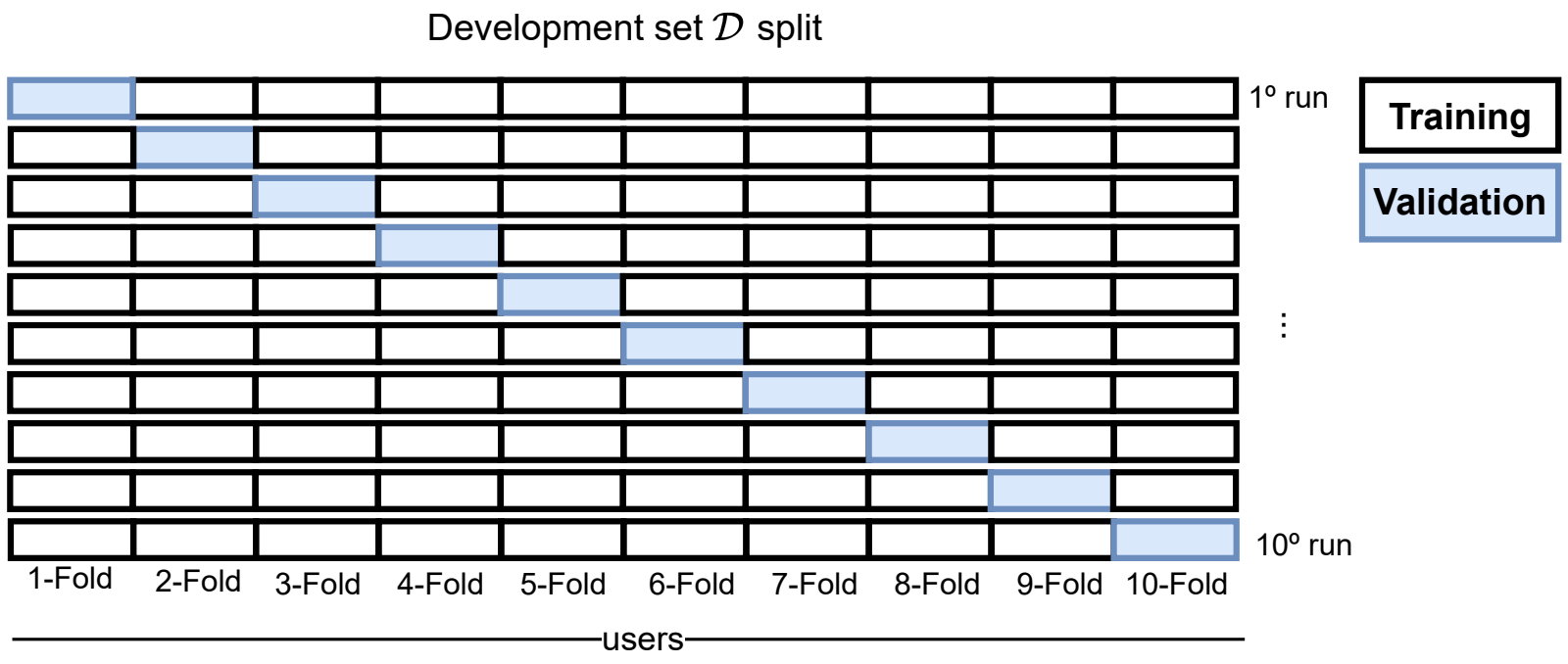


Fig. 1: Cross-validation on the development set ($\mathcal{D}$) to define the best $k$ value for the data-driven summarization. Folds are created using disjoint subsets of users to ensure that samples in the training and validation sets belong to different individuals, avoiding any user overlap during evaluation.

performed on the development set of GPDS-S, and a 5-fold cross-validation is conducted on the other datasets using random disjoint sets of users for training and validation, as shown in Figure 1. The summarization procedure is performed on each training set, and dissimilarities are computed. Specifically, positive dissimilarities are calculated between twelve genuine signatures, while the negative dissimilarities are obtained from comparisons between eleven genuine signatures and six prototypical signatures per user. For the validation sets, dissimilarities are computed based as follows:

- **CEDAR** and **GPDS Synthetic**: between twelve reference signatures and both ten genuine and ten random forgery signatures per user.
- **MCYT**: between ten reference signatures and both five genuine and five random forgery signatures per user.

The cross-validation is performed using the summarization process for each $k \in \{10, 50, 100, 150, 200, 250, 300, 350, 400, 450, 500, 550, 600\}$. For CEDAR and MCYT, we limited the maximum $k$ to 500 and 400, respectively, due to their smaller number of samples.

The data-driven process generates $k$ prototypical signatures, of which six are selected to create balanced dissimilarity sets for training classifiers. The $k$ yielding the lowest mean Equal Error Rate (EER) in detecting random forgeries is selected as the optimal value for testing.

Results for the validation step are presented in Figure 2. Charts in the first row show the best $k$ value for SVM-RBF. SVM-Linear (SGD) results are presented in the second row. As observed, the most suitable number of prototypical summarized signatures is different for each dataset and classifier. Respectively for GPDS-S, CEDAR, and MCYT, $k$ was equal to 150, 10, and 50 for SVM-RBF, and 100, 10, and 100 for SVM-Linear (SGD). These numbers correspond to a small fraction of the original development sets. With these values, train-

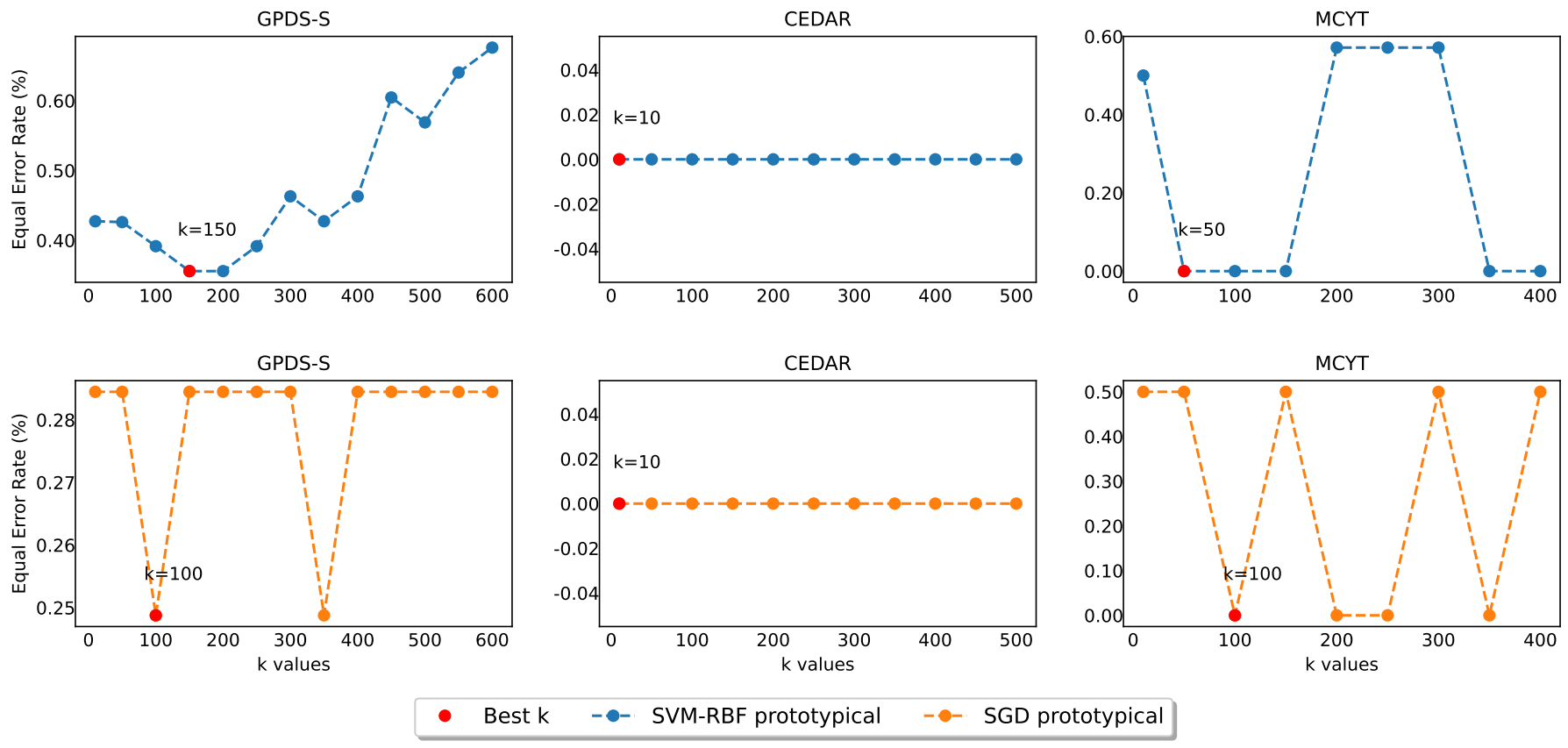


Fig. 2: WI EER for the cross-validation process. Results show random forgery detection with a global threshold for SVM-RBF and SVM-Linear (SGD).

ing data were obtained for the classifier learning process and evaluated on the exploitation sets.

## 4 Analysis of Sensitive to $k$ Value

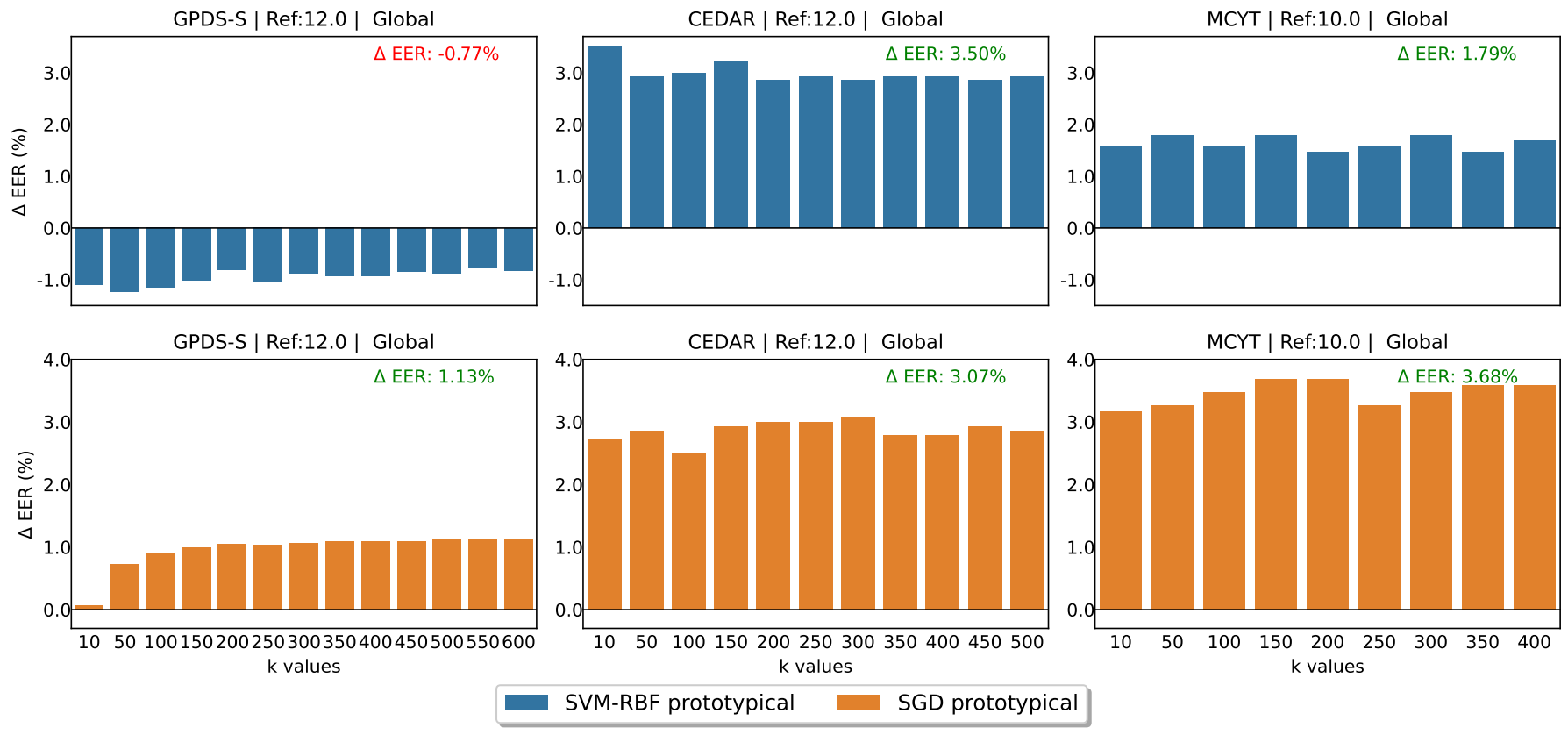


Fig. 3: Skilled forgery detection performance of the proposed method compared to the standard approach in WI settings, evaluated across different values of $k$. Testing was performed using a global threshold and the maximum number of reference signatures for SVM-RBF and SVM-Linear (SGD) classifiers. Best EER variations are marked. Green values indicate performance gains (lower EER), and red values indicate losses relative to the baseline.

As the only hyperparameter requiring explicit definition in our method, $k$ plays a central role in determining the number of prototypical signatures available for training data generation. To assess how sensitive the system's perfor-

mance is to this choice, we conducted a sensitivity analysis by evaluating system outcome across a wide range of $k$ values, from 10 to 600, on the exploitation sets.

For each dataset, we measured performance using the Equal Error Rate, and compared it against the baseline (standard approach) by computing the EER variation ($\Delta$ EER). Here, a decrease in $\Delta$ EER represents a gain in performance, while an increase indicates a loss. Results are summarized in Figure 3 for skilled forgery detection with global thresholds, using the maximum number of reference signatures. Gains are highlighted in green and losses in red.

Overall, the proposed method proves to be robust: in most cases, it either outperforms or closely matches the baseline across different $k$ values. In fact, even in the worst-case scenario for SVM-Linear (SGD) on the GPDS-S dataset, the performance drop is negligible. Meanwhile, significant gains are observed elsewhere, such as a 3.07% improvement on the CEDAR dataset.

These results suggest that the system is not overly sensitive to the choice of $k$, and maintains stable and competitive performance across a broad range of configurations. This behavior is particularly advantageous in continual learning and data stream scenarios, where explicit hyperparameter tuning is often impractical due to the absence of clear validation protocols and the evolving nature of the data [2]. In such settings, data distributions may shift over time (concept drift), and hyperparameters optimized for earlier tasks may no longer remain optimal as new data is introduced [1].

## 5 Machine Architecture Configuration

All evaluations regarding complexity measurements were conducted on a system with a 64-bit Linux operating system, with an Intel(R) Core(TM) i7-6700K processor clocked at 4.00 GHz and 32 GB of RAM. The CPU supports AVX2 and other instruction sets optimized for numerical computations.To reduce performance variability, 6 CPU cores were shielded from system interruptions using Linux CPU isolation techniques. The architecture configuration is described below:

- **Architecture**: x86_64 (Little Endian)
- **Cores/Threads**: 4 physical cores, 8 threads (1 socket, 2 threads per core)
- **L1 Cache**: 128 KiB per level (instruction and data)
- **L2 Cache**: 1 MiB (per core)
- **L3 Cache**: 8 MiB (shared)
- **Max CPU Frequency**: 4.2 GHz

## 6 Performance with Prototypical Signatures

The tabular results for prototypical signature performance are provided in Table 2 for the global threshold setting and in Table 3 for the user-specific threshold setting.

Table 2: WI average EER for skilled forgery detection across different reference ($r$) signatures using the best k values for SVM-RBF and SVM-Linear (SGD) using a global threshold. The lowest EER values are highlighted.

| Dataset | $r$ | SVM-RBF | | SVM-Linear (SGD) | |
|---|---|---|---|---|---|
| | | Standard | Prototypical | Standard | Prototypical |
| GPDS-S | 1.0 | 15.45 ± 0.59 | 15.95 ± 0.57 | 15.17 ± 0.52 | **14.79 ± 0.62** |
| | 2.0 | 12.82 ± 0.43 | 13.46 ± 0.15 | 12.67 ± 0.33 | **11.85 ± 0.21** |
| | 3.0 | 11.27 ± 0.50 | 12.10 ± 0.32 | 11.32 ± 0.57 | **10.86 ± 0.26** |
| | 5.0 | 10.17 ± 0.57 | 11.11 ± 0.22 | 10.21 ± 0.51 | **9.54 ± 0.09** |
| | 10.0 | 9.17 ± 0.37 | 10.09 ± 0.19 | 9.45 ± 0.41 | **8.33 ± 0.08** |
| | 12.0 | 8.81 ± 0.39 | 9.81 ± 0.19 | 9.09 ± 0.42 | **8.19 ± 0.06** |
| CEDAR | 1.0 | 19.57 ± 1.20 | **19.00 ± 1.20** | 19.93 ± 0.81 | 19.07 ± 1.57 |
| | 2.0 | 18.29 ± 1.14 | **16.50 ± 1.72** | 18.29 ± 0.96 | 16.71 ± 1.02 |
| | 3.0 | 16.14 ± 0.96 | **14.36 ± 0.64** | 16.14 ± 1.22 | 14.86 ± 0.86 |
| | 5.0 | 15.00 ± 0.87 | **11.79 ± 0.94** | 15.00 ± 0.56 | 12.07 ± 0.69 |
| | 10.0 | 13.29 ± 1.05 | **10.07 ± 0.47** | 13.00 ± 1.57 | 10.21 ± 0.41 |
| | 12.0 | 12.86 ± 0.80 | **9.36 ± 0.30** | 13.00 ± 1.17 | 10.29 ± 0.69 |
| MCYT | 1.0 | 19.05 ± 1.14 | **18.32 ± 3.01** | 21.16 ± 1.64 | 18.32 ± 2.25 |
| | 2.0 | 16.63 ± 1.81 | **14.84 ± 2.21** | 18.00 ± 1.95 | 15.05 ± 2.02 |
| | 3.0 | 14.53 ± 1.92 | **13.37 ± 1.99** | 15.68 ± 2.09 | 13.68 ± 1.58 |
| | 5.0 | 11.68 ± 2.15 | 10.53 ± 1.49 | 13.37 ± 1.81 | **10.32 ± 1.52** |
| | 10.0 | 9.26 ± 0.29 | **7.47 ± 1.20** | 11.05 ± 0.98 | 7.58 ± 1.09 |

Table 3: WI average EER for skilled forgery detection across different reference ($r$) signatures using the best k values for SVM-RBF and SVM-Linear (SGD) using a user-specific threshold. The lowest EER values are highlighted.

| Dataset | $r$ | SVM-RBF | | SVM-Linear (SGD) | |
|---|---|---|---|---|---|
| | | Standard | Prototypical | Standard | Prototypical |
| GPDS-S | 1.0 | 9.91 ± 0.27 | **9.77 ± 0.97** | 9.77 ± 0.42 | 10.17 ± 0.91 |
| | 2.0 | 7.77 ± 0.18 | 7.76 ± 0.51 | **7.59 ± 0.17** | 7.74 ± 0.54 |
| | 3.0 | 6.73 ± 0.47 | **6.63 ± 0.67** | 6.71 ± 0.41 | 6.87 ± 0.54 |
| | 5.0 | 5.70 ± 0.36 | 5.57 ± 0.35 | **5.57 ± 0.36** | 5.84 ± 0.24 |
| | 10.0 | 4.89 ± 0.35 | 4.75 ± 0.24 | 4.81 ± 0.40 | **4.70 ± 0.26** |
| | 12.0 | 4.61 ± 0.19 | **4.55 ± 0.24** | 4.61 ± 0.29 | 4.60 ± 0.26 |
| CEDAR | 1.0 | **9.43 ± 0.82** | 9.93 ± 1.42 | 9.50 ± 0.41 | 9.93 ± 1.30 |
| | 2.0 | 8.29 ± 0.64 | 8.14 ± 0.77 | **7.86 ± 0.25** | 8.43 ± 0.60 |
| | 3.0 | 7.50 ± 0.84 | **6.93 ± 0.96** | 7.29 ± 0.90 | 7.00 ± 0.90 |
| | 5.0 | 6.07 ± 1.21 | **4.93 ± 1.05** | 6.00 ± 1.17 | 5.00 ± 1.13 |
| | 10.0 | 4.36 ± 0.53 | **4.00 ± 0.47** | 4.50 ± 0.70 | 4.14 ± 0.82 |
| | 12.0 | 4.57 ± 0.69 | **3.79 ± 0.82** | 4.64 ± 1.04 | 3.93 ± 0.84 |
| MCYT | 1.0 | 13.89 ± 2.22 | 12.53 ± 1.88 | 14.53 ± 2.06 | **12.42 ± 1.47** |
| | 2.0 | 11.16 ± 1.46 | **10.11 ± 2.74** | 12.42 ± 1.09 | 10.42 ± 2.72 |
| | 3.0 | 9.37 ± 1.41 | **8.95 ± 2.44** | 10.95 ± 2.34 | 8.95 ± 2.66 |
| | 5.0 | 7.26 ± 1.20 | 7.16 ± 1.69 | 8.53 ± 1.68 | **6.42 ± 1.51** |
| | 10.0 | 5.05 ± 2.12 | **4.42 ± 1.42** | 6.53 ± 1.61 | 4.63 ± 1.14 |

## 7 Evaluation with Different Backbones

This section presents the validation results for selecting the $k$ value for each backbone and the corresponding global-threshold performance.

Figure 4 shows the validation curves and the selected $k$ for each backbone–dataset pair, following the procedure described in Section 3 of this supplementary material.

Using the selected $k$ for each configuration, we evaluated two classifiers on the test sets: an SVM-RBF trained with standard negative sampling, and a Linear SVM optimized with SGD using prototypical signatures. Their performance is compared in Table 4 with a global threshold.

Shapiro–Wilk tests to verify residual normality are presented in Table 5.

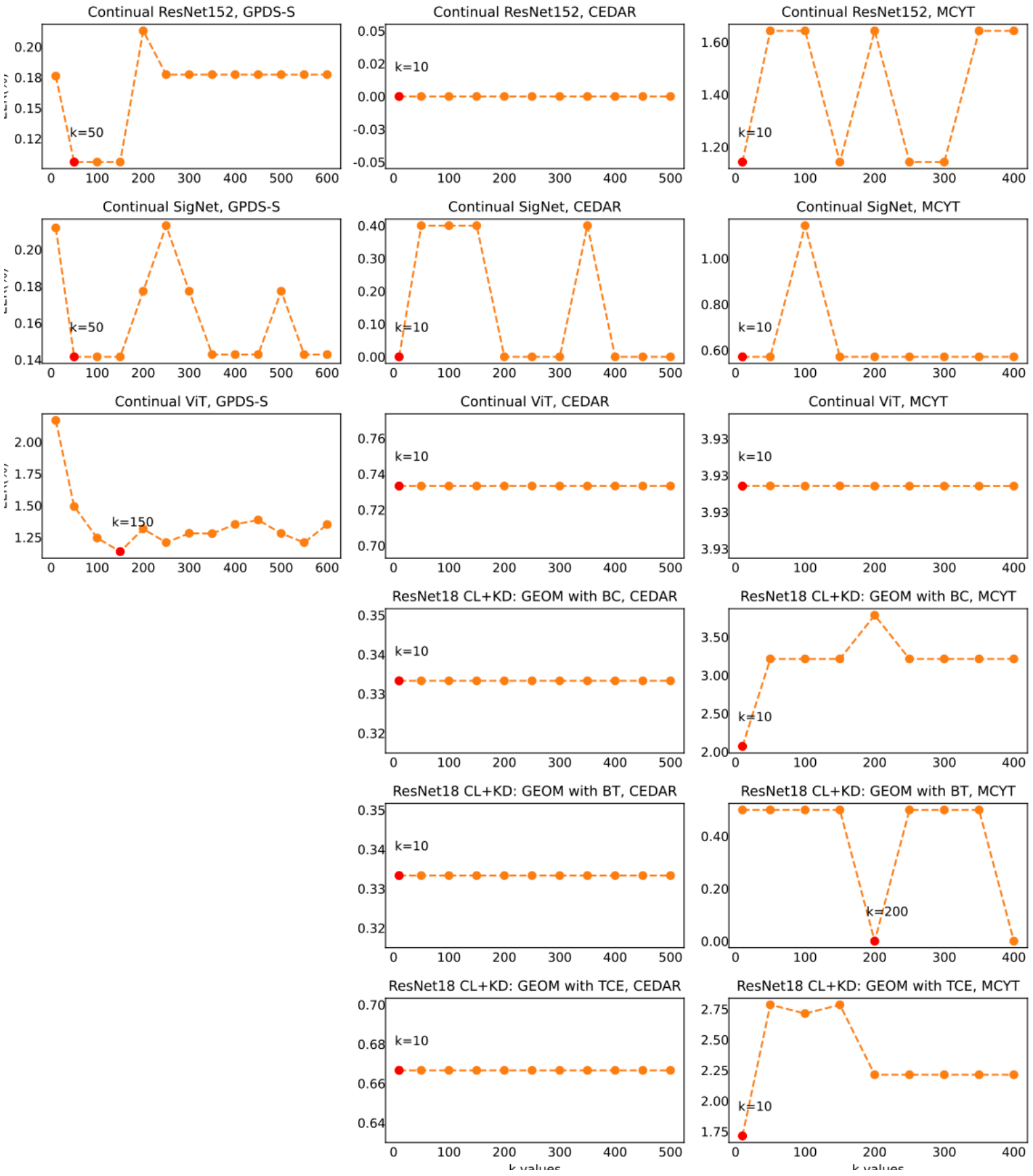


Fig. 4: WI EER for the cross-validation process for different backbone architectures. Results show random forgery detection with a global threshold for SVM-Linear (SGD).


\end{document}